\documentclass[10pt,twocolumn,letterpaper]{article}

\usepackage[pagenumbers]{cvpr} 

\usepackage{multirow}
\usepackage{bm}
\usepackage[accsupp]{axessibility}  
\definecolor{cvprblue}{rgb}{0.21,0.49,0.74}
\usepackage[pagebackref,breaklinks,colorlinks,allcolors=cvprblue]{hyperref}

\def\paperID{19} 
\def\confName{CVPR}
\def\confYear{2025}

\title{SemanticSugarBeets: A Multi-Task Framework and Dataset for Inspecting Harvest and Storage Characteristics of Sugar Beets}

\author{Gerardus Croonen ~~ Andreas Trondl ~~ Julia Simon ~~ Daniel Steininger \\
AIT Austrian Institute of Technology \\ 
Center for Vision, Automation \& Control \\
{\tt\small\{gerardus.croonen,andreas.trondl.fl,julia.simon,daniel.steininger\}@ait.ac.at}}

\begin{document}
\maketitle

\begin{abstract}
While sugar beets are stored prior to processing, they lose sugar due to factors such as microorganisms present in adherent soil and excess vegetation. Their automated visual inspection promises to aide in quality assurance and thereby increase efficiency throughout the processing chain of sugar production. In this work, we present a novel high-quality annotated dataset and two-stage method for the detection, semantic segmentation and mass estimation of post-harvest and post-storage sugar beets in monocular RGB images. We conduct extensive ablation experiments for the detection of sugar beets and their fine-grained semantic segmentation regarding damages, rot, soil adhesion and excess vegetation. For these tasks, we evaluate multiple image sizes, model architectures and encoders, as well as the influence of environmental conditions. Our experiments show an \textit{mAP\textsuperscript{50-95}} of 98.8 for sugar-beet detection and an mIoU of 64.0 for the best-performing segmentation model.
\end{abstract}

\section{Introduction}
Automated visual inspection of post-harvest sugar beets can aide in quality control by ensuring their optimal storability, thus preserving their sugar content in storage. Harvest time is short and production capacity is limited, such that large amounts of sugar beets await processing in clamps on fields or paved outdoor areas. While in storage, sugar is consumed by continued beet metabolism and microorganisms in the soil-filled crevices of sugar beets that thrive on damaged beet surfaces and residual plant material. The main cause of sugar-beet damage is the harvesting process, during which vegetation is cut from the beets (`topping'). A mismatch between beet size and cutting height can result in `under-topping' (leaving plant stems and leaves) or `over-topping' (exposing an over-sized beet interior). Additional damages, including root breakage and cracks, are incurred during mechanical handling, soil removal and transportation.

An automated inspection of sugar beets (hereafter referred to as `beets' for brevity) can prevent excessive sugar loss in storage. Online assessment of topping quality allows for the real-time adaptation of topping height. Pre-storage and in-storage quantification of soil and damages can optimize clamp processing order by prioritizing rapidly deteriorating batches of beets. Automated quantification of soil adhesion in beet deliveries can reduce the factory-side charge for soil and vegetation. Automated visual inspection can furthermore aide researchers in determining the influence of storage measures (treatments, clamp coverings and wind breakers) or help breeders evaluate the influence of genetic properties of beet varieties on their storability \cite{HoSc16}.

\begin{figure}
    \centering
    \includegraphics[width=1\columnwidth]{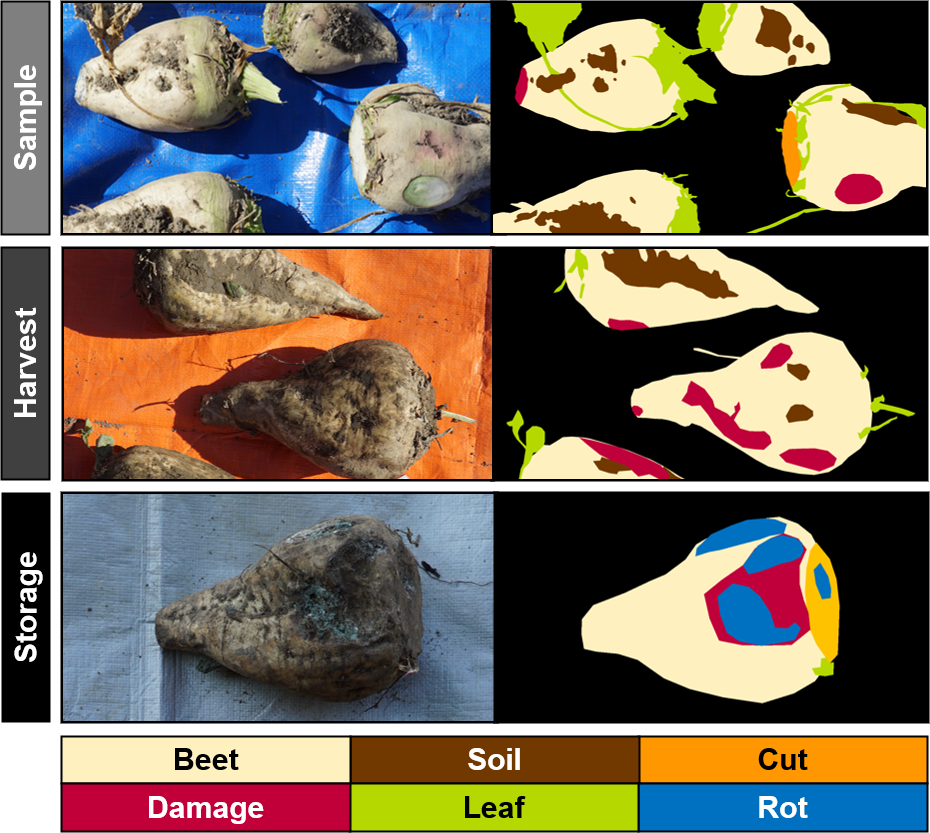}
   \caption{Representative images and semantic segmentation masks of the SemanticSugarBeets dataset from three processing stages. Label categories are listed in the color bar. The background class is displayed in black.}
    \label{fig:annotation}
\end{figure}

\subsection{Work}
The overall goal of our work is to develop and evaluate a vision-based approach to automated quality assessment of post-harvest and post-storage beets. Achieving this goal presents several challenges. First, a vision-based approach is complicated by the high variability in natural beet appearance and outdoor lighting conditions. Second,  damages and rot are relatively rare and typically occur in relatively small beet areas, leading to high class imbalance. Finally, the lack of publicly available data relevant to the problem poses a major challenge. To address these challenges, we present the following contributions:

\begin{itemize}
    \item a novel multi-task dataset of images and label masks of post-harvest and post-storage beets, including an analysis and statistics demonstrating its variability.\footnote{Dataset, code and models are available for academic use at \href{https://github.com/semanticsugarbeets/semanticsugarbeets}{https://github.com/semanticsugarbeets/semanticsugarbeets}}
    \item a two-stage method, combining fast detection with fine-grained semantic segmentation and mass estimation of beets in monocular RGB images.
    \item thorough experimentation and evaluation, showing the applicability of the work for beet quality assessment under multiple environmental conditions.
\end{itemize}
\section{Related work}
In this section, we review prior work in the field of precision agriculture relating to the proposed work of vision-based inspection of beets.

A vision-based plant row detection system using the Hough-transform on grayscale images of beets was presented in \cite{BaWo08}. In \cite{MiLo17} color and NIR images from airborne camera systems are used for vegetation-based index segmentation and custom CNN-based classification of beet plants and weeds. In \cite{MoIm21} images from airborne multi-spectral cameras are used to develop and evaluate a VGG16-based CNN for the classification of common beet-culture weeds. In \cite{JaMo06} pixel-based discriminant analysis in various color spaces is used to segment color images of beet plants and weeds under varying light conditions. In \cite{GaFr20} a YOLOv3-based method for detecting hedge bindweed, was proposed and trained on 2,271 synthetic and 472 real images of beet plants, showing high detection rates and low inference times. In \cite{LoHo16} a method for the detection of beets and weeds based on random forest classification and Markov random field smoothing on extracted image features is proposed. In \cite{NaOm22} a U-Net model, employing ResNet50 as an encoder, was trained using 1,385 color images with the aim of segmenting crop and weed images. In \cite{HaNe18} a support vector machine is used to classify image features, such as color, intensity and gradient, extracted from images of diseased beets to identify various leaf spot diseases. An Updated Faster R-CNN architecture was proposed to detect the presence and severity of leaf spot disease and trained and tested on 155 color images in \cite{OzMe19}. In \cite{AdKe23} another method for the detection of leaf spot disease, based on common CNNs, is proposed. In \cite{StHe07} it was shown that hyper-spectral leaf reflectance and multi-spectral canopy reflectance can be used to detect Beet necrotic yellow vein virus in beet plants. In \cite{MaSt10} a hyper-spectral sensor system is used to develop spectral vegetation indices, a combination of which was then used to detect the early stages of various fungal pathogens in beet plants. In \cite{PaZh15} color and NIR images of beets and beet slices were used to develop models for sucrose content prediction, showing strong correlation between certain wavelengths and sucrose content.

There are several popular publicly available extensive plant image datasets with annotations suitable for tasks from detection to semantic and instance segmentation. In most of these datasets, beet plants are a class among many. WeedAI \cite{HaZi24} provides a searchable dataset of images of different plants and their bounding boxes. Sugar Beets 2016 \cite{ChLo17} is a large-scale RGB-D and multi-spectral image dataset, capturing various growth stages of beet plants and weeds. Aerial images and annotations in the SPAGRI-AI dataset \cite{JoMu24} are suitable for plant detection tasks. CropAndWeed \cite{StTr23} provides annotations for instance segmentation of over 112,000 individual plants of over 70 different species of crop and weeds. Additionally, some datasets focus on classification of visible plant diseases, such as CottonLeafDisease \cite{BiNi24} and PlantVillage  \cite{MoHu16}.

Some available datasets display a strong focus on beet plants: LincolnBeet \cite{SaAd22} is a dataset containing images and annotations of beet plants and weeds suitable for detection, but not segmentation. The Deep Nutrient Deficiency for Sugar Beet (DND-SB) Dataset \cite{YiKr20}) provides images of nutrient deficient and wilted beet plants suitable for detection. PhenoBench \cite{WeMa24} is a dataset with a strong focus on beet plants, providing highly detailed segmentation masks for the segmentation of beets, beet instances and individual leaves. Only a few datasets specialize in beets themselves instead of the plants. In \cite{NaWi21} various CNN-based detectors, such as YOLOv4 and CSPDarknet53, are trained for the detection of mechanical damage (with cracks, breakage and surface abrasion being mapped to a single damage class) to beets in harvesters, showing good results.

To the best of our knowledge, the proposed dataset is the first of its kind in terms of size, image resolution and annotation granularity. Additionally, the proposed work is the first to introduce a two-stage approach to detection and segmentation of beets.
\section{SemanticSugarBeets dataset}
As a basis for our learning experiments, we recorded and annotated an extensive image dataset of beets. This section provides details of the acquisition and annotation process.

\subsection{Data acquisition and meta-annotation}
To cover variations in beet appearance, over the course of two years, we have recorded more than 5,000 images in ten separate recording sessions at five different locations. Beets were captured in three distinct processing stages: \textit{Sample} (harvesting, topping and simulated damages performed manually), \textit{Harvest} (automated harvesting using a harvester) and \textit{Storage} (beets from the \textit{Harvest} stage after about 90 days of storage). The beets from \textit{Sample} and \textit{Harvest} were photographed at four different agricultural fields, while the \textit{Storage} beets were captured outside the cold storage of a sugar factory. In a separate recording session, beets of the \textit{Sample} stage were weighed to allow for experimentation with beet-mass estimation.

\begin{figure}
    \centering
    \includegraphics[width=1\columnwidth]{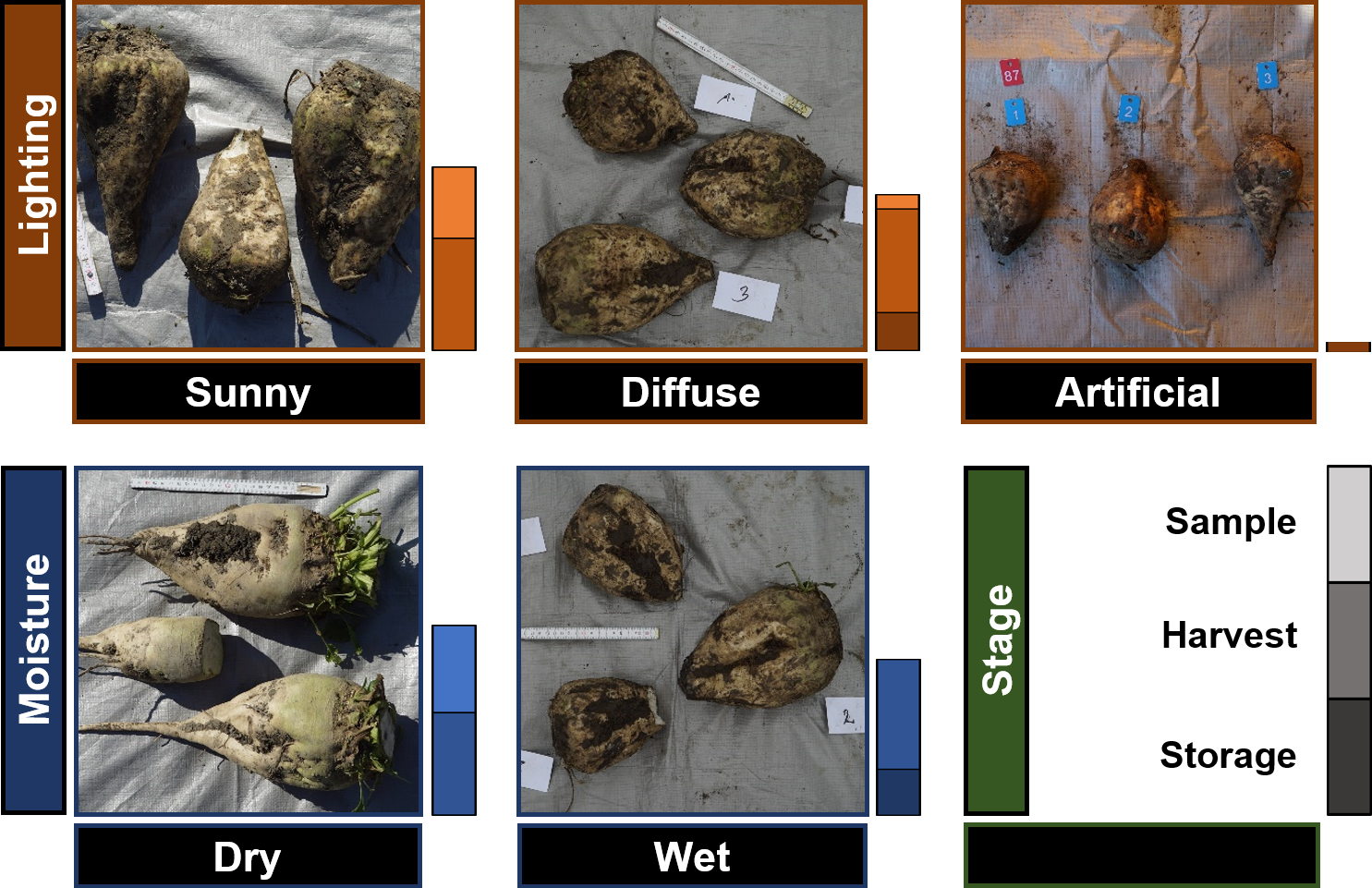}
    \caption{Representative samples of annotated lighting and soil-moisture conditions as well as their distributions of beets across the three processing stages.}
    \label{fig:meta_parameters}
\end{figure}

All images were captured with a full-frame Sony A7S camera at a resolution of 4240x2384 pixels using the following protocol: Depending on beet size, sets of two to five beets are arranged on one of four tarps of distinct colors. To determine absolute scale, a reference object (ruler or number label) of known size is placed alongside them. Beets are photographed twice to ensure correct focus and then carefully flipped and photographed twice again. Relevant meta-data such as the cultivation location, lighting conditions, degree of soil moisture and processing stage is recorded at the start of each image acquisition session. After session completion, the best of the two available images for each beet side is marked in a data sheet and incorporated into the dataset. A visualization combining representative samples for all meta-parameters and their distributions in the proposed dataset is shown in \cref{fig:meta_parameters}.

\subsection{Beet appearance interpretation}
The beets from each of the three stages \textit{Sample}, \textit{Harvest} and \textit{Storage} show distinct appearances, mostly due to lighting and soil conditions (\cref{fig:annotation}). In dry soil and late-summer weather (bright directional light), as was the case when the \textit{Sample} beets were manually harvested, beet and damage surfaces are clean, vegetation is green and adherent soil is dry. In contrast, beets in the \textit{Harvest} stage were harvested from moist soil in mid-October (diffuse low-intensity light), such that beet and damage surfaces are covered in a thin film of brownish soil and vegetation has mostly withered and turned a light brown color. Soil adhesion appears moist and fills beet crevices and indentations. Beets from the \textit{Storage} stage, after having been stored in a cooling cell for about 3 months, were photographed in a semi-indoor setting using a mixture of diffuse low-light natural and directional artificial light. The beets show dry intact beet surfaces with moist rotting areas as well as mold and fungi in varying colors (white, blue, yellow). In this stage, adherent soil is very hard to visually distinguish from the beet, even in person.

\subsection{Class definition and annotation}
Annotations are created in-house using the open-source tool \textit{Scalabel} \cite{Sc24}, and consist of manually drawn polygonal contours for each area of interest. Each area is assigned one of seven class labels relevant to the purpose of quality inspection, as visualized in \cref{fig:annotation}. The following class labels were defined for corresponding areas: \textit{Beet} - healthy, undamaged beet, \textit{Dmg} - any kind of beet damage, including cracks, except for the beet topping areas. \textit{Cut} - areas (typically one per beet) resulting from the topping process. \textit{Leaf} - fresh or dried vegetation. \textit{Soil} - residual soil adhering to beets. \textit{Rot} - areas indicating rot. \textit{Bg} - background including all remaining areas that do not qualify for any of the other labels. In addition to the semantic beet masks, in each image, the location of one of two classes of reference objects, namely rulers (\textit{Ruler}) and numbered plastic cards (\textit{Sign}) was annotated by marking its four corner points. 

To allow for beet-mass estimation from a measured beet area, a separate set of beet images and their individual mass was recorded. A subset of these images was annotated with beet and reference-marker contours. After completion of the manual annotation process using the defined classes, we implement an automatic annotation conversion to facilitate the task of instance segmentation. The conversion merges labeled polygons for selected classes of the same beet into a single polygon, assigning it the label \textit{Beet}.

\begin{figure*}
    \centering
    \includegraphics[width=2.0\columnwidth]{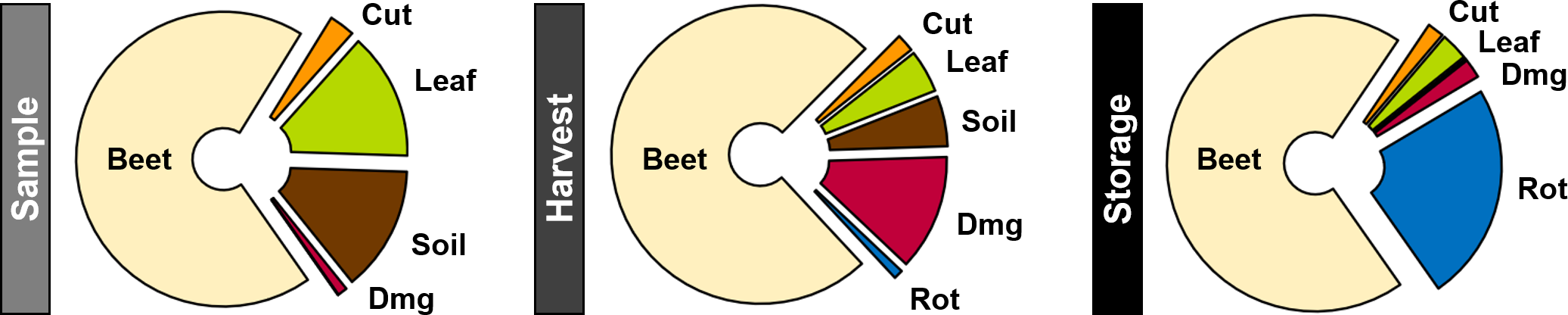}
    \caption{Distributions of annotated pixels per processing stage for \textit{Beet}, \textit{Cut}, \textit{Leaf}, \textit{Soil}, \textit{Damage} and \textit{Rot} classes.}
    \label{fig:label_distributions}
\end{figure*}

\subsection{Dataset statistics}
Overall, in 953 images selected from all ten recording sessions, a total of almost 3,000 individual beets and their components were fully annotated. Information regarding the distribution of beet parameters in images of all three defined stages is summarized in \cref{tab:dataset_statistics]}. The majority of beet images and instances is of the \textit{Sample} and \textit{Harvest} stage (86.3\%). However, the dataset also includes a relevant number of \textit{Storage} recordings with fewer beets per image. The label distributions across all stages are shown in \cref{fig:label_distributions}. They confirm that our dataset plausibly mirrors real-world conditions: As mechanical harvesting is a rather violent process, the aim of which is the removal of vegetation and soil, \textit{Harvest} beets display lower levels of soil adhesion and vegetation compared to manually harvested \textit{Sample} beets, at the cost of a higher occurrence of damage. Beet rot rarely occurs before harvesting and does not become apparent until after prolonged periods of storage.

Furthermore, a total of 1,949 reference markers of known size were annotated in 1,317 images. 58\% of these objects are folding-ruler elements and the remaining 42\% are numbered plastic cards.

\begin{table}
    \centering
        \begin{tabular}{l|crrrr|r}
            \textbf{Stage} & \textbf{Loc} & \textbf{Rec} & \textbf{Img} & \textbf{Beets} & \textbf{B/I} & \textbf{Ratio} \\
            \hline
            Sample & A & 5 & 209 & 717 & 3.4 & 24.6 \\
            Harvest & B\text{\textbar}C\text{\textbar}D & 3 & 601 & 1803 & 3.0 & 61.7 \\
            Storage & E & 2 & 143 & 400 & 2.8 & 13.7 \\
            \hline
             & \textbf{5} &  \textbf{10} & \textbf{953} & \textbf{2920} & \textbf{3.1}
        \end{tabular}
        \caption{Dataset statistics for individual \textbf{Stage}s including capturing \textbf{Loc}ations, numbers of \textbf{Rec}ording sessions, \textbf{Im}a\textbf{g}es and \textbf{Beets}, as well as average \textbf{B}eets per \textbf{I}mage and \textbf{Ratio}s of overall beets in percent.}
        \label{tab:dataset_statistics]}
\end{table}
\section{Methodology}
\begin{figure*}
    \centering
    \includegraphics[width=2.07\columnwidth]{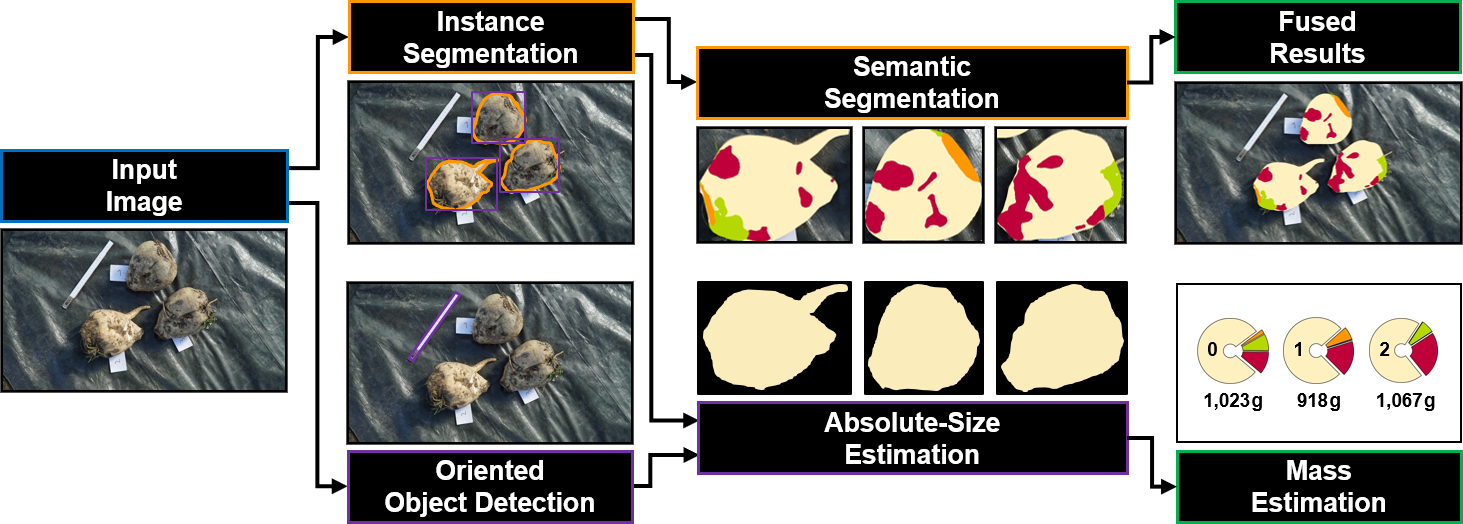}
    \caption{Overview of our two-stage approach to detection and segmentation of sugar beets. First, beets are isolated through instance segmentation. Second, detected beet patches are finely segmented. Finally, segmentation results are fused to provide semantic segmentation of vegetation, soil adhesion, damage and cutting surfaces. Additionally, oriented bounding boxes of reference markers are detected and used for inferring the mass of each beet.}
    \label{fig:processing_pipeline}
\end{figure*}

To benefit from a fast detection of individual beets as well as fine-grained semantic segmentation, our proposed approach to visual inspection of beets is a two-stage method.
As a first stage, instance segmentation and oriented object detection are performed to detect and delineate entire beets and reference objects, respectively. The measurement of a reference object's size from its detection allows for absolute size estimation of beets in the same image and subsequent mass estimation. The output of the first stage by itself can be used for further tasks such as instance counting. For our purpose, we pass the detected beet instances into a second stage to perform the more challenging task of fine-grained semantic segmentation. A comprehensive visualization depicting the processing pipeline of our proposed method is visible in \cref{fig:processing_pipeline}.

The proposed approach relies on the results of three separate learning tasks: instance segmentation of individual beets, semantic segmentation of detected beet instances and oriented object detection of reference objects. The setup for each of these three tasks is described in detail in the following subsections.

\subsection{Learning experiments}
We generate a random dataset split \textit{once} with a ratio of 70/15/15 between training, validation and test data respectively. Additionally, we ensure that images displaying different sides of the same beet are in the same set. The resulting dataset split is then used across all learning experiments for both beets and reference objects, ensuring a completely independent test set for all learning tasks throughout the entire processing pipeline.

\paragraph{Coarse-grained instance segmentation}
We use our automatic conversion method to merge all types of beet areas corresponding to the same instance into a single area labeled \textit{Beet}, while assigning \textit{Leaf} to \textit{Bg} as vegetation, especially in case of under-topped beets, can overlap with other beets and distort their area-based mass estimation. The resulting dataset after conversion is a one-class dataset in which each beet is combined into a single instance. We train our models using the instance-segmentation module of YOLO11 with adaptive batch size and default hyper-parameters on an NVIDIA RTX 2080 and repeat the experiment for multiple combinations of model variants and input sizes. The best model is selected based on accuracy and inference time after 400 epochs.

\paragraph{Fine-grained semantic segmentation}
To determine an optimal balance between performance, inference time and model size, we train and evaluate multiple network architectures and encoders for input image sizes \textit{Large} (1056x576), \textit{Medium} (768x448) and \textit{Small} (512x288). To conduct our experiments, we use PyTorch and \textit{pytorch-segmentation-models} \cite{IaPa19}, which supports a large number of architectures and encoders. However, not every combination of architectures and encoders results in a functioning model. We choose PSPNet \cite{ZhSh17}, MANet \cite{LiZh21} and U-Net \cite{RoFi15} as architectures for their compatibility with all of the following encoders: EfficientNet (B0) \cite{TaLe19}, MIT (B0) \cite{DoBr15}, MobileNetV3 (Large 100) \cite{HoZh19}, MobileOne (S0) \cite{ZhLi21}, RegNetY (002) \cite{DzRo20}. We use ImageNet \cite{DeDo09} weights to initialize the encoders in all our experiments. All hyper-parameters are kept at their default value, unless otherwise specified.

We determine the maximum batch size for the largest image size by training for one epoch on an NVIDIA RTX 3090 with 24 GB of VRAM and found it to be 4. For reasons of comparability between all architecture and encoder combinations, we use this batch size for all experiments. We use a one-cycle learning rate scheduler and set the maximum learning rate to $10^{-3}$. We select AdamW \cite{LoHu17} as the optimizer and, to deal with the high level of class imbalance in the dataset, we use Dice-loss \cite{SuLi17} (Equation \ref{eq:dice_loss}) as a loss function. In total, given 3 image sizes, 3 architectures and 5 encoders, we conduct 45 experiments. Each individual experiment is run on the fixed data split as previously discussed for a maximum of 100 epochs, stopping training early as soon as the loss does not decrease for 5 epochs.

\begin{equation}
\mathcal{L}_{\text{Dice}} = 1 - \frac{2 \sum_i p_i g_i + \epsilon}{\sum_i p_i + \sum_i g_i + \epsilon}
\label{eq:dice_loss}
\end{equation}

\paragraph{Marker detection and mass estimation}
Our experiments with oriented object detection of reference markers, i.e. folding-ruler elements and plastic signs, are performed using YOLO11 with default hyper-parameters and adaptive batch size on an NVIDIA RTX 2080 with 12 GB of VRAM. We train and evaluate four combinations of model variants and input sizes.

For mass estimation, we rely on the subset of separately annotated beets of known mass. As a pre-processing step, we calculate the depicted areas of these beets in $pixel^2$ from their segmentation masks and infer their absolute areas in $mm^2$ from the known size of the annotated reference objects. We derive the average mass per unit area $\bar{m}$ in $g/mm^2$. At runtime, the mass of detected beets can then be approximated by multiplying their absolute area by $\bar{m}$.

\subsection{Evaluation protocol and metrics}
All reported performance scores are calculated on the test set common to all learning tasks, which is completely withheld during training in all experiments.

To determine instance-segmentation and oriented-object-detection performance, we report average precision (AP) and mAP\textsuperscript{50-95} as defined by the MS COCO benchmark \cite{LiMa14} for both axis-aligned bounding boxes and instance contours of detected beets, as well as oriented bounding boxes of reference markers.

For semantic segmentation, we determine the Intersection over Union (IoU) per class as well as the mIoU \cite{EvEs15} averaged over all classes (Equation \ref{eq:miou}).

\begin{equation}
    mIoU = \frac{1}{n} \sum_{c=1}^{n}\frac{TP_c}{TP_c + FP_c + FN_c}
    \label{eq:miou}
\end{equation}

where \(TP_c\), \(FP_c\) and \(FN_c\) denote the numbers of true positive, false positive and false negative pixels, respectively, for class \(c\), and \(n\) is the total number of classes.
\section{Experiments}
The next sections describe experiment results followed by a discussion of meta-parameter influence on performance.

\subsection{Coarse-grained instance segmentation}
The results of beet instance-segmentation experiments using two YOLO11 model variants are listed in \cref{tab:iseg_results}. They show high performance across all model and input sizes. Masks are detected slightly more accurately than bounding boxes. The high performance achieved in this task is expected, as the beets are spatially isolated with limited variation in background appearance. We select the smaller model with an input size of 896 pixels for further experiments since it achieves performance similar to the large model at a reduced runtime.

\begin{table}
    \centering
    \begin{tabular}{c|c|cc|c}
        \multicolumn{1}{r}{\textbf{}} & \textbf{} & \textbf{Box} & \textbf{Mask} & \textbf{t} \\
        \hline
        \multirow{2}{*}{\textbf{s}} & \textbf{640} & 96.4 & 98.2 & 2.7 \\
         & \textbf{896} & \textbf{96.6} & \textbf{98.8} & 5.1 \\
         \hline
        \multirow{2}{*}{\textbf{l}} & \textbf{640} & 96.2 & 98.4 & 8.2 \\
         & \textbf{896} & 96.3 & 98.7 & 16.0
    \end{tabular}
    \caption{Coarse instance segmentation results as \textit{mAP\textsuperscript{50-95}} on test set for bounding \textbf{Box}es and segmentation \textbf{Mask}s. All values are listed for the model capacities \textit{Small} (\textbf{s}) and \textit{Large} (\textbf{l}) as well as input sizes of \textbf{640} and \textbf{896} pixels, along with inference \textbf{t}ime.}
    \label{tab:iseg_results}
\end{table}

\subsection{Fine-grained semantic segmentation}
\begin{figure}
    \centering
    \includegraphics[width=1\columnwidth]{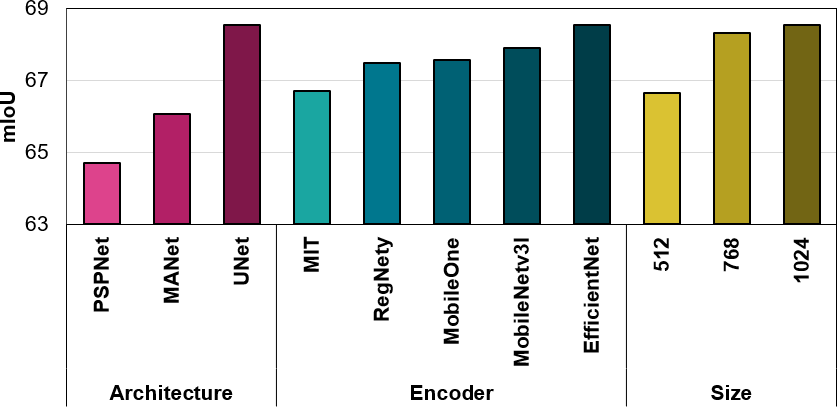}
    \caption{Maximum semantic-segmentation performance for each architecture, encoder and input image size.}
    \label{fig:eval_seg}
\end{figure}

\begin{figure}
    \centering
    \includegraphics[width=1\columnwidth]{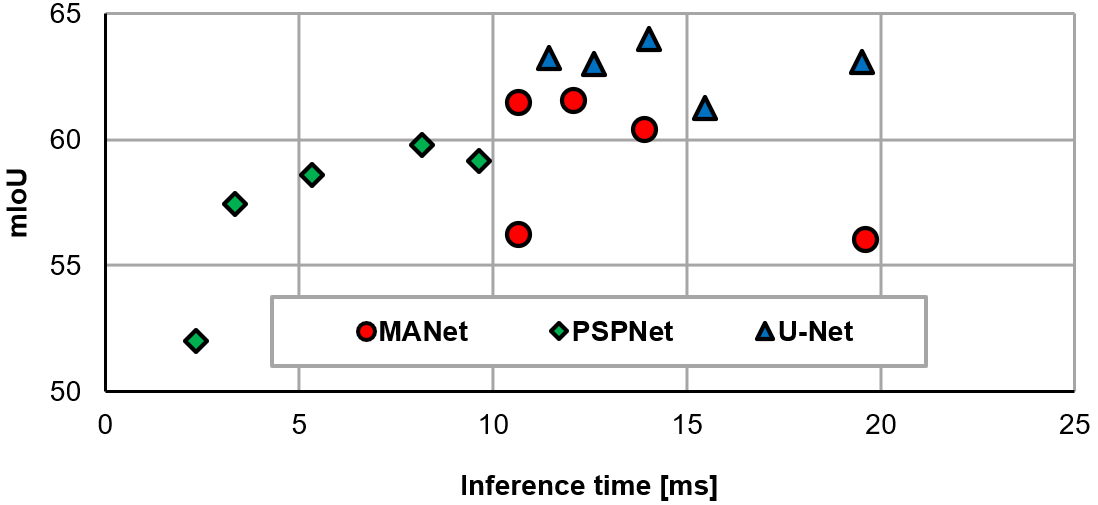}
    \caption{Scatter plot of segmentation performance (vertical axis) vs. inference time (horizontal axis) on \textit{Large} image patches.}
    \label{fig:seg_miou_vs_inf}
\end{figure}

For the semantic-segmentation task, the maximum performances for all evaluated architectures, encoders and input sizes are summarized in \cref{fig:eval_seg}. Regarding the average, mIoUs are marginally higher for \textit{Large} image patches (64.6) as compared to \textit{Medium} (64.4) and \textit{Small} image patches (62.9). Detailed results by architecture, encoder and class are presented for \textit{Large} image patches in \cref{tab:seg_results} and corresponding inference times are visualized in \cref{fig:seg_miou_vs_inf}.

\begin{table*}
    \centering
    \begin{tabular}{rl|cccccccccrr}
        \multicolumn{1}{l}{\textbf{Architecture}} & \multicolumn{1}{l|}{\textbf{Encoder}} & \multicolumn{1}{c}{\textbf{Bg}} & \multicolumn{1}{c}{\textbf{Beet}} & \multicolumn{1}{c}{\textbf{Cut}} & \multicolumn{1}{c}{\textbf{Leaf}} & \multicolumn{1}{c}{\textbf{Soil}} & \multicolumn{1}{c}{\textbf{Dmg}} & \multicolumn{1}{c|}{\textbf{Rot}} & \multicolumn{2}{c|}{\textbf{Mean}} & \multicolumn{2}{c}{\textbf{t\textsubscript{1024}}} \\
        \hline
        & \multicolumn{1}{l|}{EfficientNet} & 94.7 & 82.0 & 61.1 & 46.0 & 59.2 & 33.7 & \multicolumn{1}{l|}{54.2} & 61.6 & \multicolumn{1}{l|}{} & 12.1 &  \\
        & \multicolumn{1}{l|}{MIT} & 94.8 & 81.8 & 54.0 & 46.2 & 59.2 & 33.6 & \multicolumn{1}{l|}{53.4} & 60.4 & \multicolumn{1}{l|}{} & 13.9 &  \\
        MANet & \multicolumn{1}{l|}{MobileOne} & 93.8 & 76.8 & 55.1 & 39.7 & 53.1 & 26.1 & \multicolumn{1}{l|}{47.6} & 56.0 & \multicolumn{1}{l|}{59.2} & 19.6 & 13.4 \\
        & \multicolumn{1}{l|}{MobileNetV3} & 94.6 & 82.0 & 59.5 & 45.3 & 58.3 & 0.0 & \multicolumn{1}{l|}{54.0} & 56.2 & \multicolumn{1}{l|}{} & \textbf{10.6} &  \\
        & \multicolumn{1}{l|}{RegNetY} & 94.7 & 81.4 & 62.0 & 46.0 & 58.9 & 32.6 & \multicolumn{1}{l|}{54.9} & 61.5 & \multicolumn{1}{l|}{} & \textbf{10.6} &  \\
        \hline
        & \multicolumn{1}{l|}{EfficientNet} & 93.9 & 79.9 & 57.1 & 40.6 & 56.2 & 30.2 & \multicolumn{1}{l|}{52.3} & 58.6 & \multicolumn{1}{l|}{} & 5.3 &  \\
        & \multicolumn{1}{l|}{MIT} & 94.2 & 80.5 & 58.1 & 41.5 & 55.4 & 29.6 & \multicolumn{1}{l|}{54.7} & 59.1 & \multicolumn{1}{l|}{} & 9.6 &  \\
        PSPNet & \multicolumn{1}{l|}{MobileOne} & 94.3 & 81.2 & 56.9 & 42.6 & 57.2 & 31.1 & \multicolumn{1}{l|}{55.3} & \textbf{59.8} & \multicolumn{1}{l|}{57.4} & 8.2 & \textbf{5.8} \\
        & \multicolumn{1}{l|}{MobileNetV3} & 93.7 & 78.6 & 55.5 & 40.0 & 54.7 & 29.1 & \multicolumn{1}{l|}{50.6} & 57.4 & \multicolumn{1}{l|}{} & 3.3 &  \\
        & \multicolumn{1}{l|}{RegNetY} & 92.9 & 75.3 & 43.1 & 33.4 & 48.4 & 24.7 & \multicolumn{1}{l|}{46.2} & 52.0 & \multicolumn{1}{l|}{} & \textbf{2.3} &  \\
        \hline
        & \multicolumn{1}{l|}{EfficientNet} & \multicolumn{1}{c}{\textbf{95.1}} & \multicolumn{1}{c}{\textbf{84.6}} & \multicolumn{1}{c}{\textbf{63.1}} & \multicolumn{1}{c}{\textbf{49.6}} & \multicolumn{1}{c}{\textbf{61.0}} & \multicolumn{1}{c}{\textbf{36.6}} & \multicolumn{1}{c|}{58.0} & \multicolumn{1}{c}{\textbf{64.0}} & \multicolumn{1}{c|}{} & \multicolumn{1}{c}{14.0} & \multicolumn{1}{c}{} \\ & \multicolumn{1}{l|}{MIT} & \multicolumn{1}{c}{94.9} & \multicolumn{1}{c}{83.2} & \multicolumn{1}{c}{53.4} & \multicolumn{1}{c}{47.9} & \multicolumn{1}{c}{59.9} & \multicolumn{1}{c}{34.2} & \multicolumn{1}{c|}{55.3} & \multicolumn{1}{c}{61.3} & \multicolumn{1}{c|}{} & \multicolumn{1}{c}{15.5} & \multicolumn{1}{c}{} \\
        U-Net & \multicolumn{1}{l|}{MobileOne} & \multicolumn{1}{c}{94.9} & \multicolumn{1}{c}{83.3} & \multicolumn{1}{c}{61.5} & \multicolumn{1}{c}{48.5} & \multicolumn{1}{c}{60.2} & \multicolumn{1}{c}{35.3} & \multicolumn{1}{c|}{57.9} & \multicolumn{1}{c}{63.1} & \multicolumn{1}{c|}{\textbf{62.9}} & \multicolumn{1}{c}{19.5} & \multicolumn{1}{c}{14.6} \\
        & \multicolumn{1}{l|}{MobileNetV3} & \multicolumn{1}{c}{95.0} & \multicolumn{1}{c}{84.6} & \multicolumn{1}{c}{61.8} & \multicolumn{1}{c}{49.2} & \multicolumn{1}{c}{59.4} & \multicolumn{1}{c}{35.5} & \multicolumn{1}{c|}{57.2} & \multicolumn{1}{c}{63.3} & \multicolumn{1}{c|}{} & \multicolumn{1}{c}{\textbf{11.4}} & \multicolumn{1}{c}{} \\
        & \multicolumn{1}{l|}{RegNetY} & \multicolumn{1}{c}{95.0} & \multicolumn{1}{c}{83.6} & \multicolumn{1}{c}{59.9} & \multicolumn{1}{c}{48.3} & \multicolumn{1}{c}{59.3} & \multicolumn{1}{c}{35.8} & \multicolumn{1}{c|}{\textbf{58.8}} & \multicolumn{1}{c}{63.0} & \multicolumn{1}{c|}{} & \multicolumn{1}{c}{12.6} & \multicolumn{1}{c}{} \\
        \hline
        & \textbf{Mean} & 94.4 & 81.3 & 57.5 & 44.3 & 57.4 & 29.9 & 54.0 &  &  &  &
    \end{tabular}
    \caption{Results of semantic segmentation experiments on \textit{Large} test-set image patches for all combinations of three \textbf{Architecture}s and five \textbf{Encoder}s, separated by class. Performance numbers represent per-class IoU reported as a percentage. The best performing combination of architecture and encoder is highlighted in bold. The best mIoU for each architecture is highlighted in bold in the penultimate column. The rightmost column shows inference times \textbf{t} in ms.}
    \label{tab:seg_results}
\end{table*}

The results show similar average performance independent of architecture and encoder. The combination of \textit{U-Net} architecture and \textit{EfficientNet} encoder shows the best overall performance, but is slightly outperformed in the \textit{Rot} class by the combination of the same architecture with the \textit{RegNetY} encoder. On average, models of the \textit{U-Net} architecture perform best, independently of encoder. From the individual per-class performance, it is evident that all models perform best for the classes \textit{Bg} and \textit{Beet}, significantly lower for \textit{Cut}, \textit{Leaf}, \textit{Soil} and \textit{Rot} and still lower for the \textit{Damage} class. The relatively low performance for this class is most likely due to under-representation in the dataset as well as the ambiguity of this class against the \textit{Cut} class; Damaged areas near topping surfaces are sometimes classified as part of the topping surface. In terms of inference times PSPNet-based models outperform the other models by a significant amount, as visible in \cref{fig:seg_miou_vs_inf}.

\begin{figure}
    \centering
    \includegraphics[width=1\columnwidth]{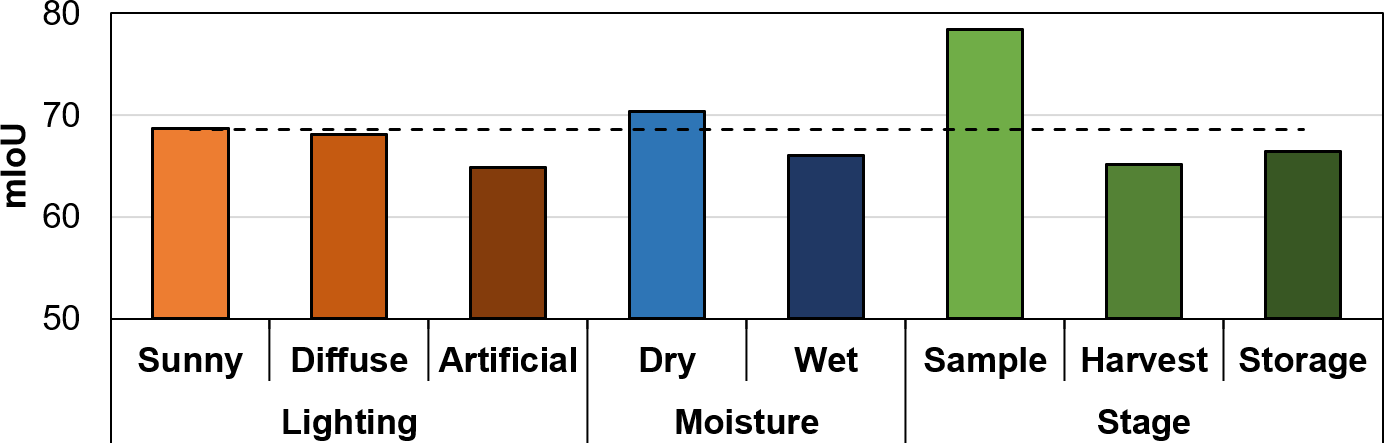}
    \caption{Fine-grained segmentation performance by meta-parameter for the best-performing model. The dashed line indicates the overall mIoU.}
    \label{fig:influence_meta_seg}
\end{figure}

Overall, the best-performing combination consists of the \textit{U-Net} architecture with the \textit{EfficientNet\-b0} encoder on \textit{Large} image patches. To measure the ability of this model to adapt to the apparent beet-appearance variations and label distributions between the different processing stages (\cref{fig:annotation,fig:label_distributions}), we measure the average mIoU per test sample for each value of each category (lighting, moisture, stage). The results are shown in \cref{fig:influence_meta_seg}.

The largest performance difference in terms of lighting is between natural (sunny, diffuse) and artificial lighting, which is not caused by lighting alone, as all \textit{Storage} beets were photographed under artificial light. Wet beets are slightly harder to segment than dry ones which is likely due to their darker appearance, causing wet beet areas and soil adhesion to appear more similar. Beets in the \textit{Sample} stage are easiest to segment, as these beets have been  harvested from dry soils, showing the most distinct appearance between classes. Beets in \textit{Storage} combine the most challenging conditions of being wet and photographed in artificial lighting. Their relatively high performance compared to beets from the \textit{Harvest} stage is likely due to very large \textit{Rot} areas present in \textit{Storage} beets.

\subsection{Marker detection}
For detecting markers, we conduct experiments for two YOLO11 oriented-object-detection model variants and two input sizes. Results for both classes of reference objects are listed in \cref{tab:marker_detection_results}. Overall, performance increases with larger image sizes, but not necessarily with larger model capacities, indicating that the target objects are sufficiently distinctive to be reliably detected by smaller architectures. Plastic labels (\textit{Sign}) are more accurately detected than the larger folding-ruler elements (\textit{Ruler}), most likely caused by their higher contrast to most backgrounds and less elongated shape. However, robustness is sufficiently high for both classes to perform accurate scale normalization. We select the best-performing model (small, input size of 896 pixels) for use in our proof-of-concept pipeline.

\begin{table}
    \centering
        \begin{tabular}{c|c|cc|c|c}
            \multicolumn{1}{r}{\textbf{}} & \textbf{} & \textbf{Ruler} & \textbf{Sign} & \textbf{Mean} & \textbf{t} \\
            \hline
            \multirow{2}{*}{\textbf{s}} & \textbf{640} & 92.5 & 99.2 & 95.9 & 2.1 \\
             & \textbf{896} & \textbf{93.0} & \textbf{99.3} & \textbf{96.2} & 4.1 \\
            \hline
            \multirow{2}{*}{\textbf{l}} & \textbf{640} & 92.5 & 99.2 & 95.9 & 6.8 \\
             & \textbf{896} & 92.8 & \textbf{99.3} & 96.1 & 13.3
        \end{tabular}
        \caption{Oriented-object-detection results as \textit{mAP\textsuperscript{50-95}} for \textbf{Ruler} and \textbf{Sign} markers as well as their \textbf{Mean} score on the test set. Values are listed for model capacities \textit{Small} (\textbf{s}) and \textit{Large} (\textbf{l}) at input sizes of \textbf{640} and \textbf{896} pixels, along with inference \textbf{t}ime.}
        \label{tab:marker_detection_results}
\end{table}

\subsection{Discussion and limitations}
\begin{figure*}
    \centering
    \includegraphics[width=2.07\columnwidth]{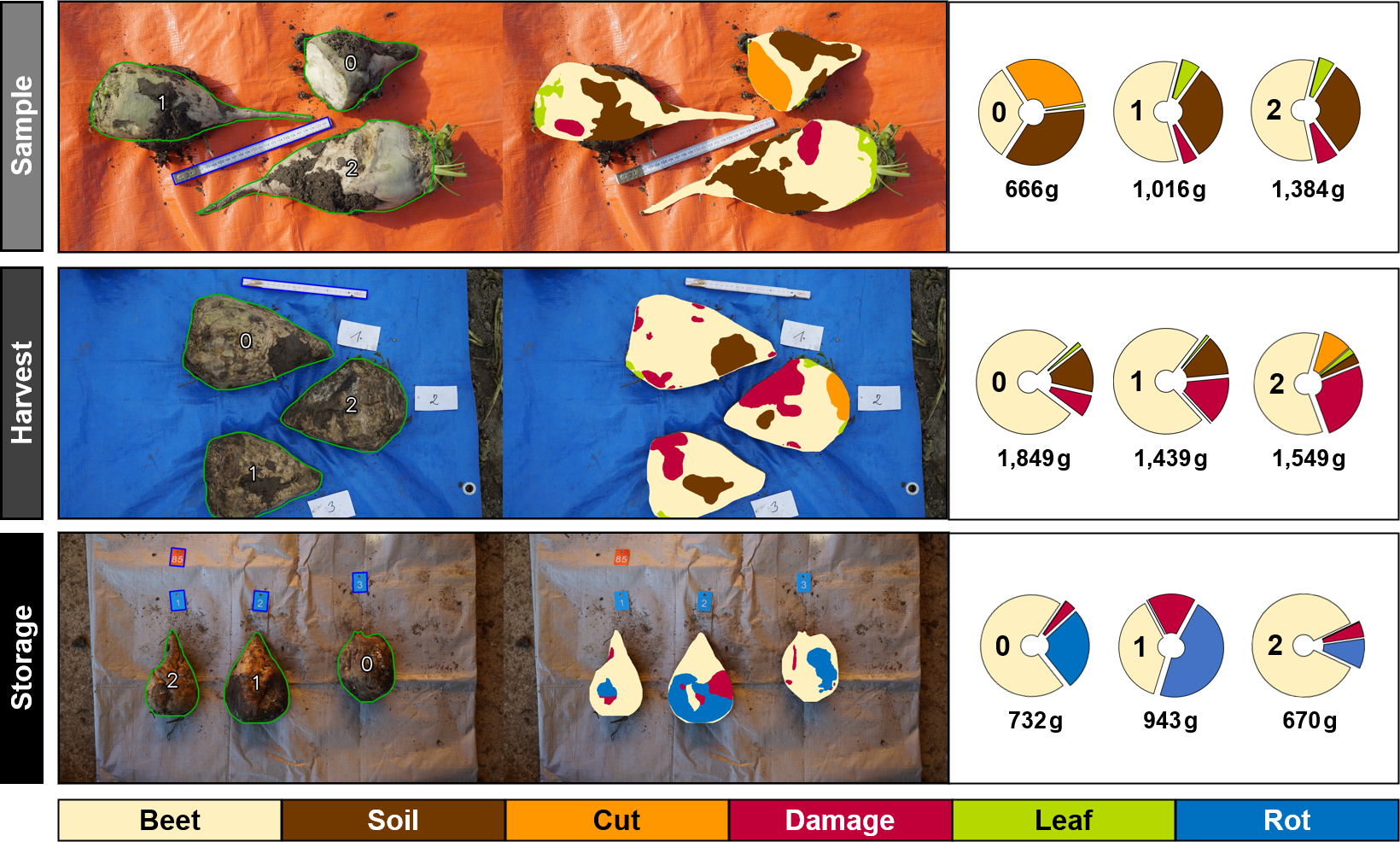}
    \caption{Representative results on test-set images for each processing stage. The leftmost images show object contours provided by coarse-grained instance-segmentation of individual beets (green), as well as oriented object detection of reference markers (blue). Fine-grained semantic segmentation within beet contours is visualized in the center column according to the color bar below. The rightmost images show per-class area distributions and mass estimates for each individual beet.}
    \label{fig:results}
\end{figure*}

Our results indicate that a two-step method for the semantic segmentation of sugar beets performs well. In most cases the first step of instance-segmentation provides highly accurate detection and delineation results for individual beets at a low run-time. The fine-grained semantic segmentation within the detected beet contours provides an accurate segmentation of individual areas of interest. Despite the varying appearance of beets between processing stages, areas are mostly assigned correctly, even differentiating between classes of similar appearance (\textit{Cut} and \textit{Damage}). Evaluating the results, we found two main performance limitations in semantic segmentation. Firstly, in rare cases, the high level of class imbalance in the dataset can cause certain classes to not be learned (MANet-MobileNetV3). Secondly, in some cases the inferred segmentation result appears more plausible than the annotation, for instance for small areas of soil adhesion and damage that are easily overlooked during annotation. The overall distributions of beet areas and estimated beet masses are found to be plausible throughout our qualitative evaluation. \cref{fig:results} visualizes the results of our entire processing pipeline.

\section{Conclusion}
The proposed work provides three main contributions: a fully annotated image dataset of post-harvest and post-storage sugar beets, a generic two-stage method for their detection, semantic segmentation as well as mass estimation, and the results of thorough experimentation, including the influence of environmental conditions. Our findings suggest that light-weight YOLO11 models are suitable for the online instance segmentation of sugar-beet contours and oriented object detection of scale reference objects. Segmenting the bounding boxes of beet contours with a trained U-Net model and any of the evaluated encoders provides accurate segmentation relevant to beet quality control.

\subsection{Outlook}
Although our dataset captures sufficient variability for real-world applications, it would benefit from covering even more environmental conditions by the addition of corresponding images. Class imbalance can be further reduced by adding more images showing damage and rot. Adding classes and images for mold and fungi could further increase the application range of the dataset.

While the results of our proof-of-concept test cases are promising, more work is needed to perform detection and segmentation in scenarios with more realistic and relevant backgrounds, such as soil, concrete or asphalt surfaces. This could be achieved by adding relevant images and annotations or by replacing image backgrounds with real or synthetic ones. The method could furthermore be extended to cover scenarios showing higher overlap between instances, such as beets on conveyor belts or in clamps, as well. Although the presented area-based method for beet-mass estimation produces plausible results, methods that estimate beet volume directly could produce more accurate estimates, but also increase complexity.

We expect the proposed method and framework to be able to adapt to further scenarios beyond the scope of this work that involve different appearances of sugar beets. Similarly, the proposed method and framework can be extended to datasets of other agricultural crops such as carrots or potatoes. By publicly providing our dataset and framework, we hope to further develop them in cooperation with the research community.

\paragraph{Acknowledgements}
The authors would like to express their gratitude to Marion Seiter, Martina Dokal, Wibke Imgenberg, Gereon Heller and Herbert Eigner of the AGRANA Research and Innovation Center (ARIC) for their instrumental contributions to this research.

{
    \small
    \bibliographystyle{ieeenat_fullname}
    \bibliography{main}
}

\newpage
\setcounter{page}{1}
\twocolumn[
     \centering
     \Large
     \textbf{SemanticSugarBeets: A Multi-Task Framework and Dataset for Inspecting Harvest and Storage Characteristics of Sugar Beets \\
     \vspace{0.7em}(Supplementary Material)} \\
     \vspace{8.2em}]

\appendix
This supplementary complements the main paper with a more detailed description of the image acquisition protocol, extended dataset statistics and additional discussion of results.

\section{Image acquisition protocol}
For the purpose of repeatability and consistency, we performed the following steps during image acquisition:

\begin{enumerate}
    \item Compose a group of beets (3 to 5) to fit inside the camera frame, held in landscape mode
    \item Put a folding ruler (or other object of known size) in the frame, ensuring its full visibility
    \item From a standing position, take two (almost identical) photographs from a top-view perspective
    \item Flip the beets and put them back in roughly the same position
    \item Force a camera refocus by taking a photograph of a nearby object, such as your hand. This photo will also allow for the quick identification of separate beet groups and beet sides when viewing and meta-annotating the photos.
    \item Repeat steps 2-3.
\end{enumerate}

\section{Extended dataset analysis}
\begin{figure}
    \centering
    \includegraphics[width=0.9\columnwidth]{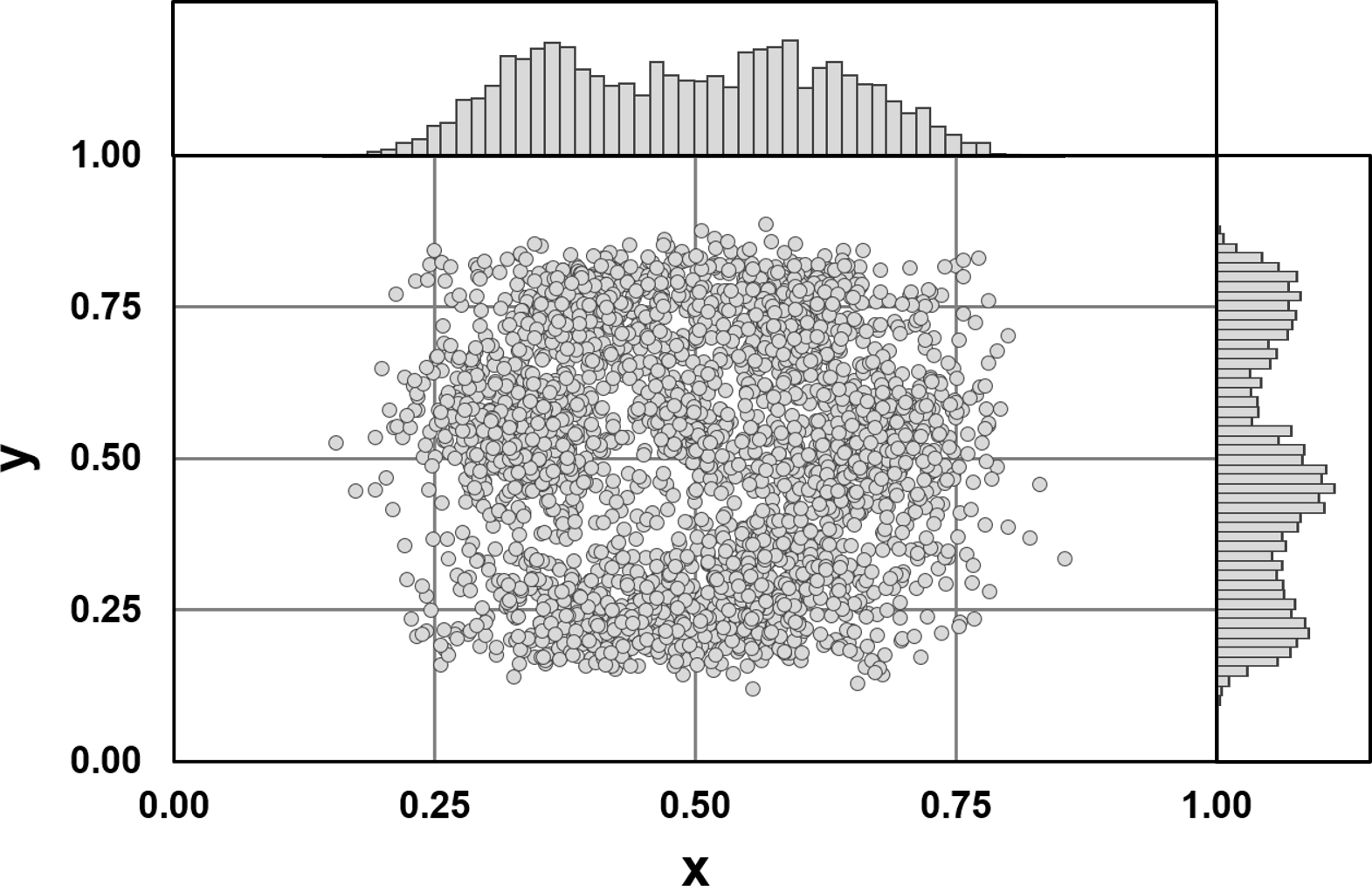}
    \caption{Distribution of normalized bounding-box centers of all annotated sugar-beet instances.}
    \label{fig:distribution_bbox_centers}
\end{figure}

\begin{figure}
    \centering
    \includegraphics[width=1.0\columnwidth]{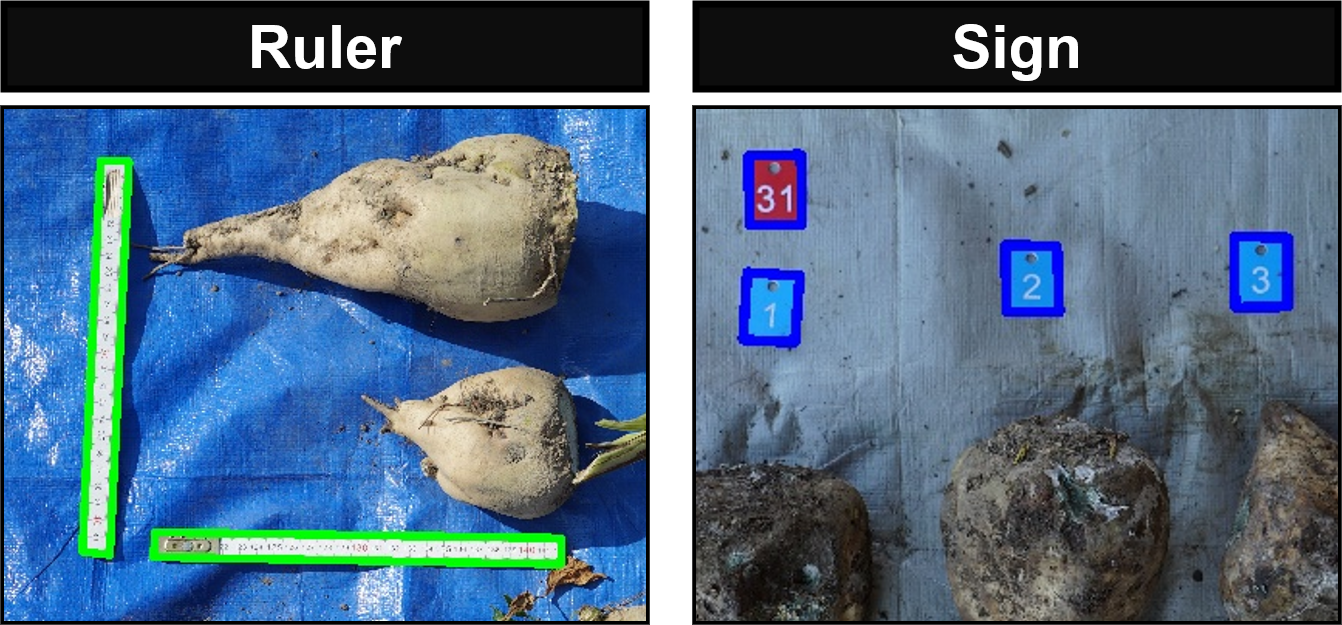}
    \caption{Representative examples of annotated reference objects.}
    \label{fig:annotation_markers}
\end{figure}

\cref{tab:dataset_acquisition_description_complete]} provides a complete list of recording sessions and corresponding statistics and meta-parameters. The distribution of bounding box centers across all beet instances is depicted in \cref{fig:distribution_bbox_centers}. Representative examples for both classes of annotated reference markers are visualized in \cref{fig:annotation_markers}.

\begin{table*}
    \begin{center}
        \begin{tabular}{c|rrc|cccc|c}
            \textbf{SID} & \textbf{Images} & \textbf{Beets} & \textbf{B/I} & \textbf{Lighting} & \textbf{Moisture} & \textbf{Marker} & \textbf{Stage} & \textbf{Loc} \\
            \hline
            0 & 33 & 165 & 5.0 & Sunny & Dry & None & Sample & A \\
            1 & 92 & 300 & 3.3 & Sunny & Dry & Ruler & Sample & A \\
            2 & 40 & 120 & 3.0 & Diffuse & Dry & Ruler & Sample & A \\
            3 & 40 & 120 & 3.0 & Sunny & Dry & Ruler & Sample & A \\
            4 & 4 & 12 & 3.0 & Sunny & Dry & Ruler & Sample & A \\
            5 & 31 & 93 & 3.0 & Sunny & Wet & Ruler & Harvest & B/C \\
            6 & 288 & 864 & 3.0 & Diffuse & Wet & Ruler & Harvest & C \\
            7 & 282 & 846 & 3.0 & Sunny & Dry & Ruler & Harvest & D \\
            8 & 116 & 319 & 2.8 & Diffuse & Wet & Sign & Storage & E \\
            9 & 27 & 81 & 2.8 & Artificial & Wet & Sign & Storage & E \\
            \hline
            & \textbf{953} & \textbf{2920} & \textbf{3.1} & & & & &
        \end{tabular}
        \caption{Parameters of recording sessions, including \textbf{S}ession \textbf{ID}, numbers of annotated \textbf{Images} and \textbf{Beets}, average ratios of \textbf{B}eets per \textbf{I}mage, \textbf{Lighting} conditions, beet \textbf{Moisture}, the presence of folding-ruler elements or plastic signs as \textbf{Marker} devices, processing \textbf{Stages} and cultivation \textbf{Loc}ations.}
        \label{tab:dataset_acquisition_description_complete]}
    \end{center}
\end{table*}

\section{Extended methodology}
\begin{figure*}
    \centering
    \includegraphics[width=2.07\columnwidth]{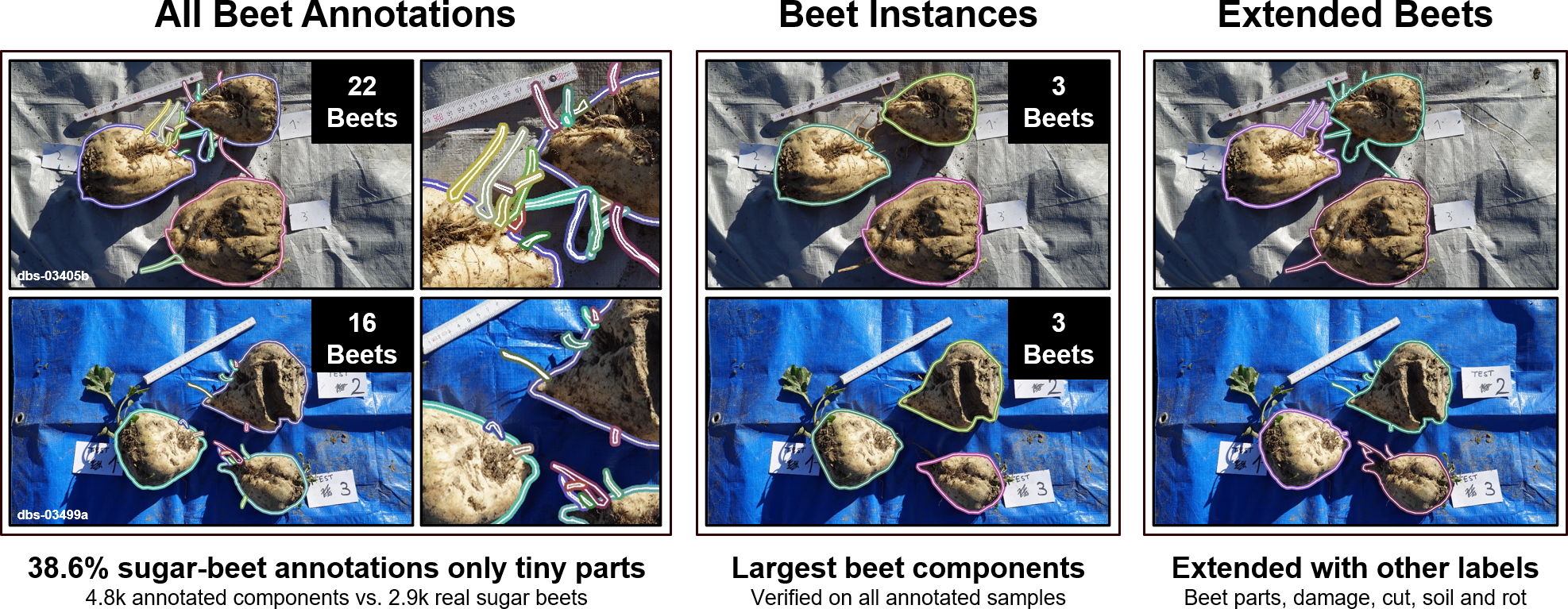}
    \caption{Annotation-synthesis pipeline to convert original annotations to instance-segmentation annotations of entire sugar beets.}
    \label{fig:annotation_synthesis}
\end{figure*}

As the original annotations usually contain multiple polygons of different fine-grained classes for a single sugar-beet, a method for automatic pre-processing is required to extract the final annotations compatible with instance segmentation. This process is visualized in \cref{fig:annotation_synthesis}. It consists of identifying the largest component of each beet and then extending it with all overlapping smaller components to derive annotations compatible with the coarse-grained instance-segmentation task described in Sec. 4.1 of the main paper.

\section{Extended evaluation results}
\cref{tab:evaluation_semseg]} provides the full evaluation details of our semantic-segmentation ablation study, a summary of which is provided in Fig. 5 of the main paper. Regarding the influence of meta-parameters on performance, the exact values forming the basis for Fig. 7 of the paper is summarized in \cref{tab:evaluation_influence_semseg]}.

\begin{table*}
    \begin{center}
        \begin{tabular}{rl|ccc|c|ccc}
            \textbf{Architecture} & \textbf{Encoder} & \textbf{512} & \textbf{768} & \textbf{1024} & \textbf{Mean} & \textbf{t\textsubscript{512}} & \textbf{t\textsubscript{768}} & \textbf{t\textsubscript{1024}} \\
            \hline
            \multirow{5}{*}{MANet} & EfficientNet & 61.2 & 62.1 & 65.4 & \multirow{5}{*}{62.8} & 9.5	& 9.9 & 12.1 \\
             & MIT & 64.0 & 64.5 & 65.6 &  & \textbf{6.2} & \textbf{7.8} & 13.9 \\
             & MobileOne & 59.5 & 61.2 & 61.3 &  & 14.4 & 16.6 & 19.6 \\
             & MobileNetV3l & 60.6 & 64.4 & 60.9 &  & 6.5 & 8.3 & \textbf{10.6} \\
             & RegNetY & 60.1 & 65.4 & 66.1 &  & 7.9 & 9.2 & \textbf{10.6} \\
             \hline
             \multirow{5}{*}{PSPNet} & EfficientNet & 62.5 & 63.0 & 63.5 & \multirow{5}{*}{62.2} & 2.6	& 3.4 & 5.3 \\
             & MIT & 63.1 & 63.9 & 63.9 & & 4.5 & 5.3 & 9.6 \\
             & MobileOne & 63.4 & 64.7 & 64.6 & & 6.1 & 6.5 & 8.2 \\
             & MobileNetV3l & 61.6 & 62.5 & 62.5 & & 2.3 & 2.6 & 3.3 \\
             & RegNetY & 58.1 & 58.0 & 57.8 &  & \textbf{1.7} & \textbf{1.9} & \textbf{2.3} \\        
             \hline
             \multirow{5}{*}{U-Net} & EfficientNet & \textbf{66.7} & \textbf{68.3} & \textbf{68.6} & \multirow{5}{*}{\textbf{67.0}} & 8.1	& 10.2 & 14.0 \\
             & MIT & 65.6 & 66.7 & 66.3 &  & \textbf{7.0}	& 8.7 & 15.5 \\
             & MobileOne & 65.8 & 66.7 & 67.6 &  & 15.3 & 16.1 & 19.5\\
             & MobileNetV3l & 66.0 & 67.9 & 67.9 &  & 7.1 & \textbf{8.3} & \textbf{11.4} \\
             & RegNetY & 65.9 & 66.9 & 67.5 & & \textbf{7.0} & 8.9 & 12.6 \\
             \hline
             & \textbf{Mean} & \textbf{62.9} & \textbf{64.4} & \textbf{64.6} & & & & \\
        \end{tabular}
        \caption{Results of semantic segmentation on the test set including all combinations of architectures, encoders and image sizes. Performance numbers are mIoUs, reported in percentages. The rightmost three columns show the inference time for each image size in ms.}
        \label{tab:evaluation_semseg]}
    \end{center}
\end{table*}

\begin{table*}
    \begin{center}
        \begin{tabular}{ll|c}
            & & \textbf{mIoU} \\
            \hline
            \multirow{3}{*}{\textbf{Lighting}} & Sunny & \textbf{68.6} \\
             & Diffuse & 68.1 \\
             & Artificial & 64.8 \\
            \hline
            \multirow{2}{*}{\textbf{Moisture}} & Dry & \textbf{70.4} \\
            & Wet & 66.0 \\
            \hline
            \multirow{3}{*}{\textbf{Stage}} & Sample & 78.4 \\
            & Harvest & 65.2 \\
            & Storage & \textbf{66.4}
        \end{tabular}
        \caption{Test set performance of semantic segmentation, separated by meta-parameter.}
        \label{tab:evaluation_influence_semseg]}
    \end{center}
\end{table*}

\end{document}